\documentclass{article}

\usepackage[utf8]{inputenc} 
\usepackage[T1]{fontenc}    
\usepackage{hyperref}       
\usepackage{url}            
\usepackage{booktabs}       
\usepackage{amsfonts}       
\usepackage{nicefrac}       
\usepackage{microtype}      
\usepackage{graphicx}
\usepackage{amsmath,amssymb} 
\usepackage{color}
\usepackage{hyphenat}
\usepackage{subfigure}
\long\def\invis#1{}
\usepackage{multirow}
\usepackage{tipa}
\usepackage{caption}

\usepackage[final,nonatbib]{nips_2018}

\usepackage[utf8]{inputenc} 
\usepackage[T1]{fontenc}    
\usepackage{hyperref}       
\usepackage{url}            
\usepackage{booktabs}       
\usepackage{amsfonts}       
\usepackage{nicefrac}       
\usepackage{microtype}      

\title{Towards Robust Lung Segmentation in Chest Radiographs with Deep Learning}

\author{
  Jyoti Islam \\
  Department of Computer Science\\
  Georgia State University\\
  \texttt{jislam2@student.gsu.edu} \\
  \And
Yanqing Zhang \\
Department of Computer Science \\
Georgia State University \\
   \texttt{yzhang@gsu.edu} \\
}

\begin{document}

\maketitle

\begin{abstract}
Automated segmentation of Lungs plays a crucial role in the computer-aided diagnosis of chest X-Ray (CXR) images. Developing an efficient Lung segmentation model is challenging because of difficulties such as the presence of several edges at the rib cage and clavicle, inconsistent lung shape among different individuals, and the appearance of the lung apex. In this paper, we propose a robust model for Lung segmentation in Chest Radiographs. Our model learns to ignore the irrelevant regions in an input Chest Radiograph while highlighting regions useful for lung segmentation. The proposed model is evaluated on two public chest X-Ray datasets (Montgomery County, MD, USA, and Shenzhen No. 3 People’s Hospital in China). The experimental result with a DICE score of 98.6\% demonstrates the robustness of our proposed lung segmentation approach.

\end{abstract}
\section{Introduction}
Chest Radiographs are the most common radiological procedure and constitute about one-third of all radiological procedures \cite{Ginneken}. Chest X-Rays (CXR) are used to study various structures such as the heart and lungs for several disease diagnosis including Lung cancer, Tuberculosis, Pneumonia etc. For computer-aided diagnostic systems, segmentation of anatomical structures in chest X-Rays plays an important role. For example, irregular shape, size measurements and total lung area can provide significant insight about early manifestations of life-threating diseases, including cardiomegaly, emphysema etc. The performance of Lung segmentation plays a vital role in such applications. Accurate Lung segmentation is considered as one of the challenges in medical image analysis due to the shape variance caused by age, gender and health. If there are a presence of external objects, such as cardiac pacemakers, surgical clips, and sternal wire, automated segmentation of Lung fields becomes more difficult.\\

There are four major categories of Lung segmentation methods. Rule-based models use predefined anatomical rules for lung segmentation \cite{Duryea}, \cite{Armato}. Pixel-based methods try to label each pixel as lung or non-lung \cite{Gin2}. Deformable-based models use object shape and image appearance \cite{Shao}. Registration based models match and refine lung fields based on a segmented lung database \cite{Candemir}. Most of the traditional models use hand-crafted shape and region information for Lung segmentation. In recent days, the advancement of deep learning technologies \cite{Litjens}, \cite{Islam_visual}, \cite{Islam2017}, \cite{Islam_journal},  \cite{Islam_cvpr} is transforming the medical image analysis world with great success . Now we can develop robust frameworks to learn useful features directly from the input data for the segmentation task. 
\begin{figure}[h]
\begin{center}
\leavevmode
\begin{tabular}{ccc}
\subfigure[]{\includegraphics[width=.25\linewidth,height=1in]{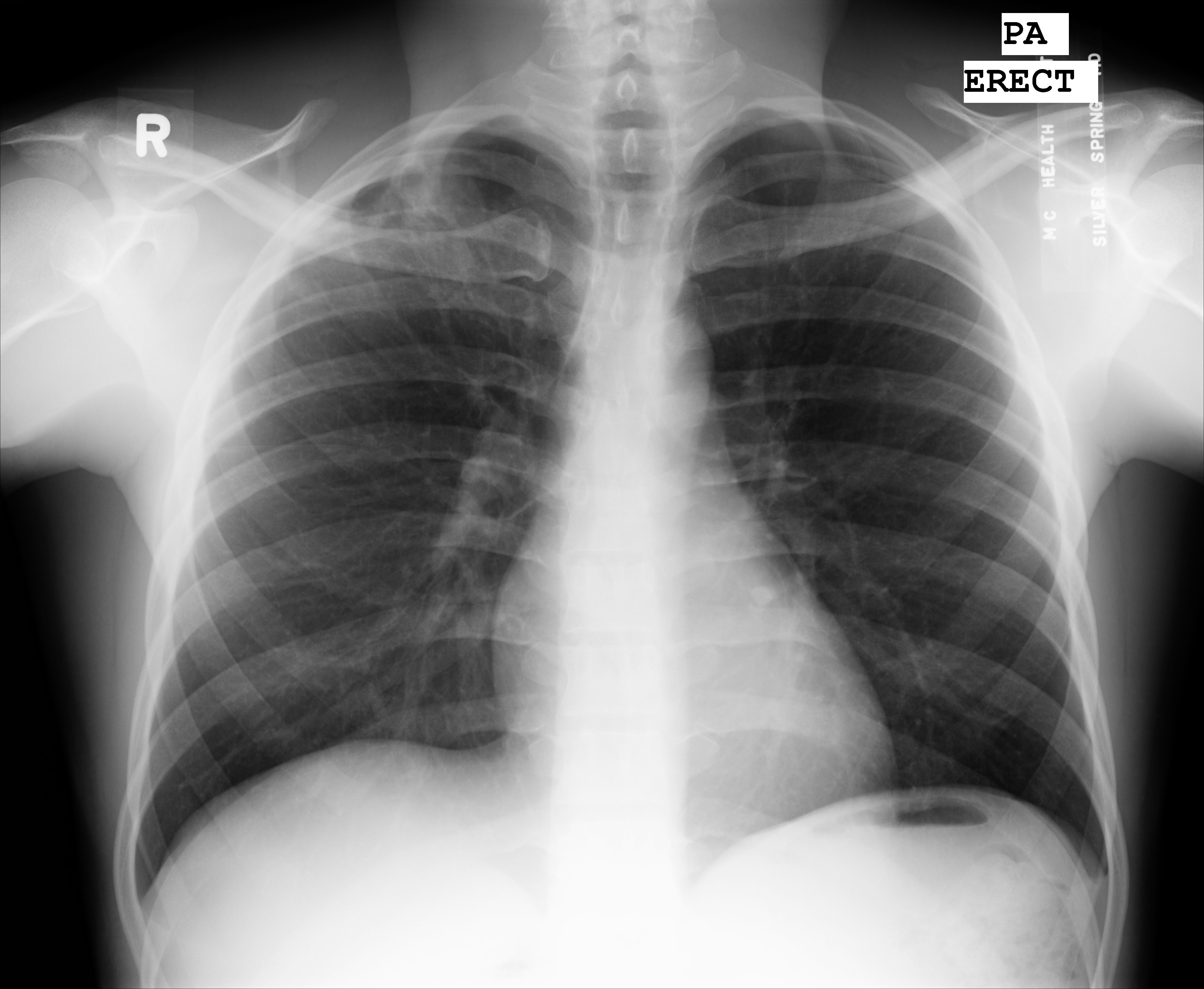}\label{fig:resp1}}&
\subfigure[]{\includegraphics[width=.25\linewidth,height=1in]{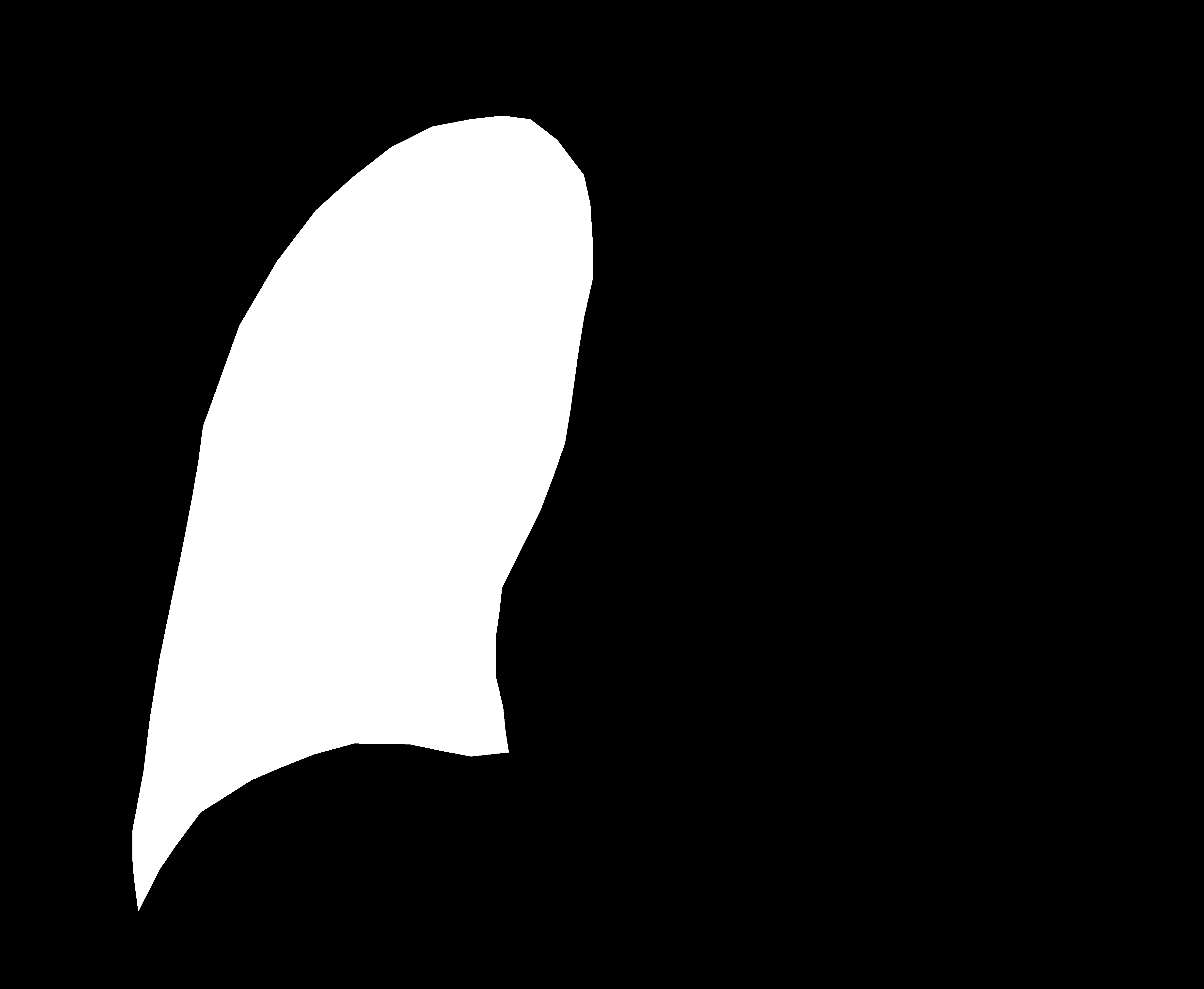}\label{fig:resp2}}&
\subfigure[]{\includegraphics[width=.25\linewidth,height=1in]{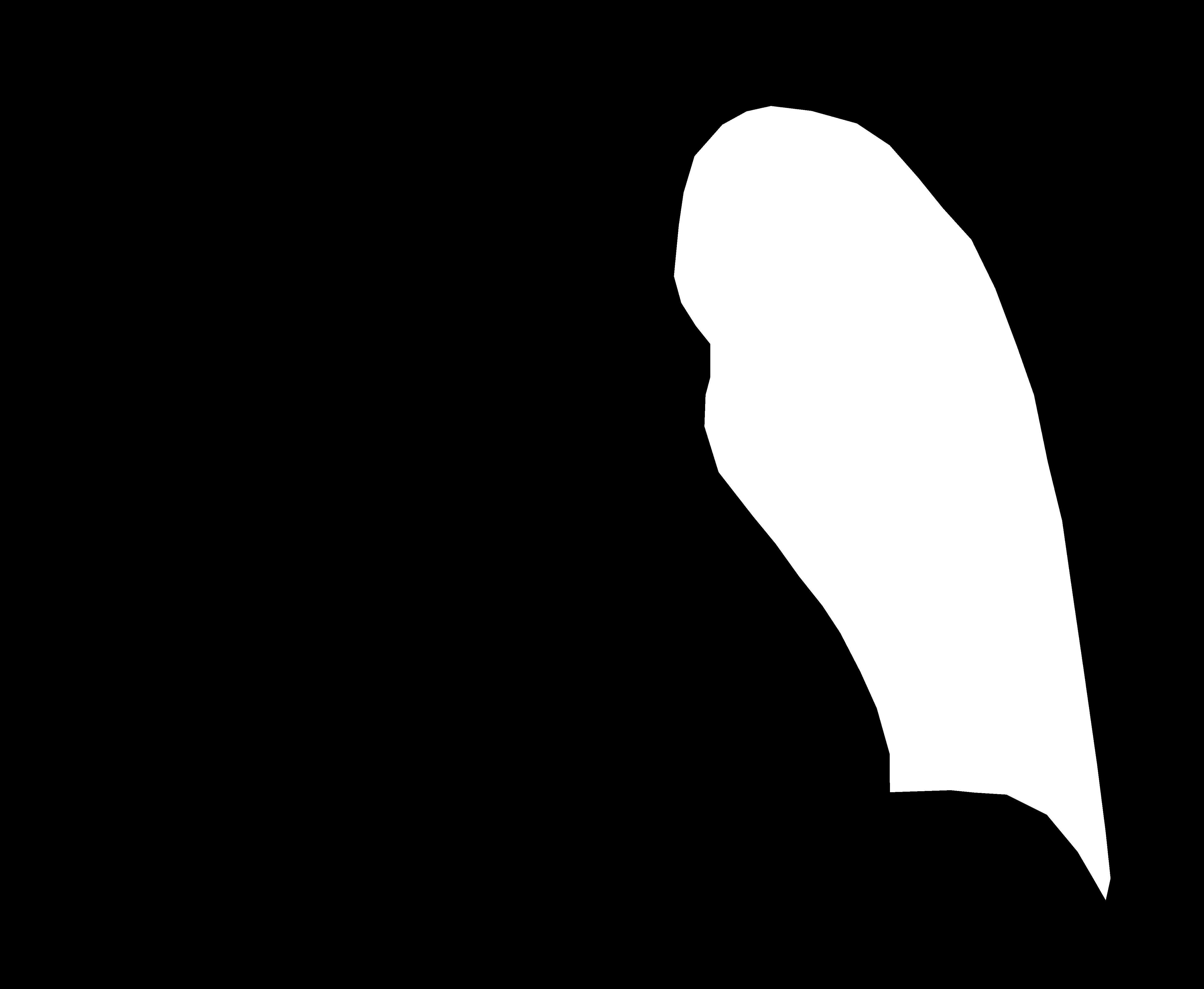}\label{fig:resp3}}
\end{tabular}
\end{center}
\caption{Sample data from Montgomery Dataset: \subref{fig:resp1} Chest X-Ray \subref{fig:resp2} Left mask; 
\subref{fig:resp3} Right mask}
\label{fig_Motgomery}
\end{figure}

For our current work, we develop an automated framework for Lung segmentation in chest X-Ray images using a Deep Convolutional Neural Network based on the U-Net \cite{unet}. Besides, with several experiments, we demonstrate that proper data augmentation and network architecture can significantly improve the performance for lung segmentation in chest X-Ray images.

\begin{figure}[h]
\begin{center}
\leavevmode
\begin{tabular}{ccc}
\subfigure[]{\includegraphics[width=.30\linewidth,height=1in]{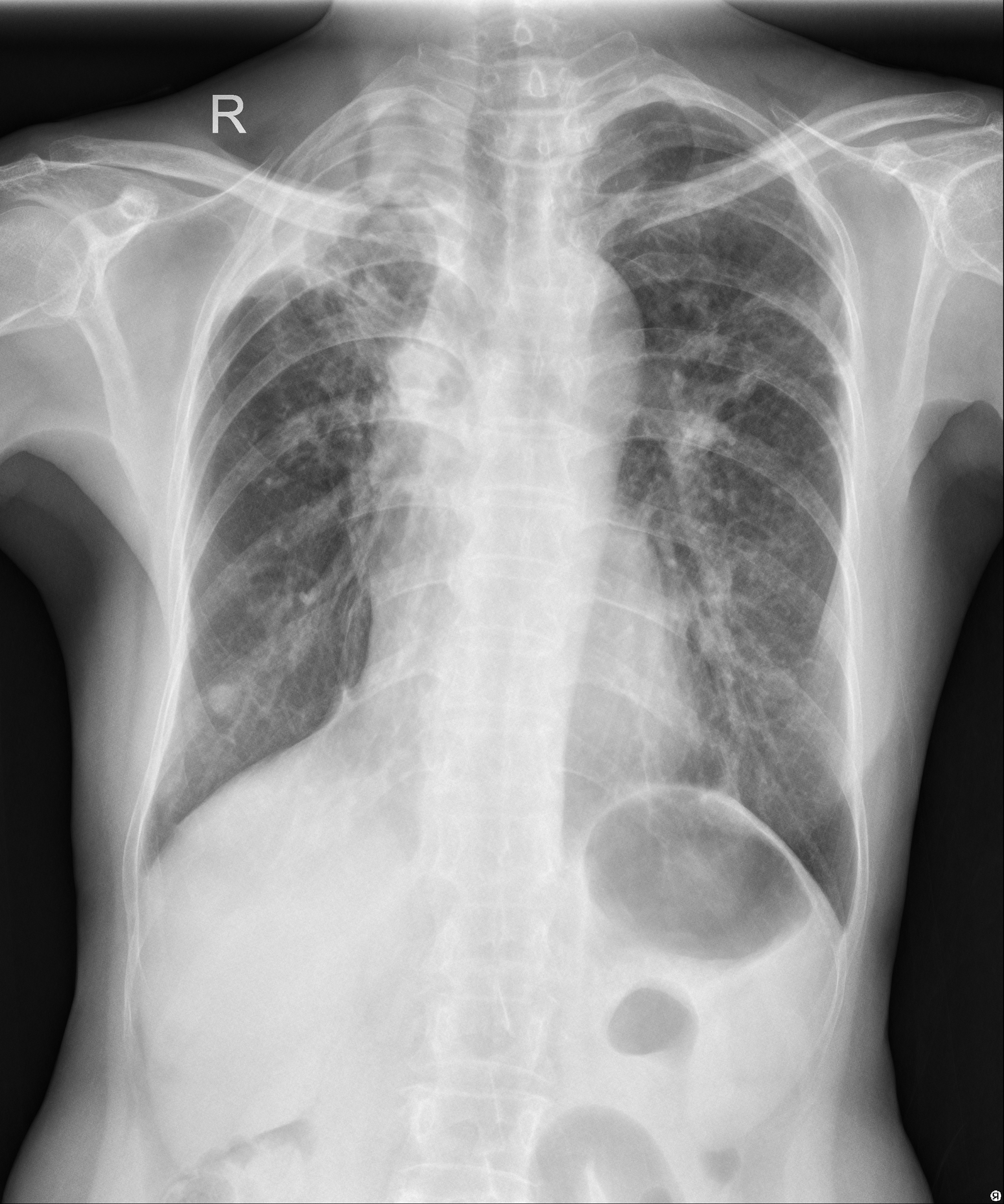}\label{fig:resp4}}&
\subfigure[]{\includegraphics[width=.30\linewidth,height=1in]{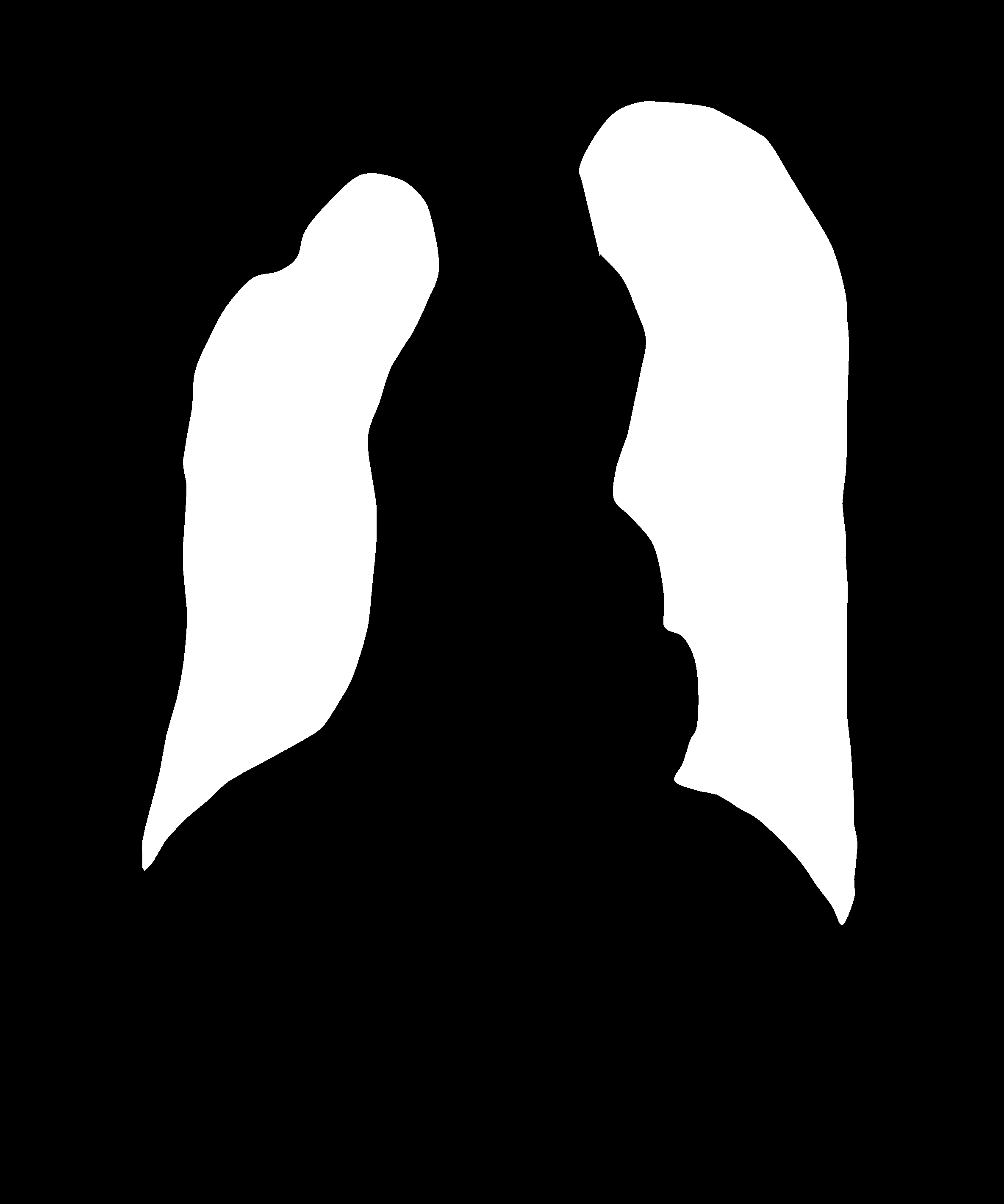}\label{fig:resp5}}& \\
\end{tabular}
\end{center}
\caption{Sample data from Shenzhen Dataset: \subref{fig:resp4} Chest X-Ray \subref{fig:resp5} mask}
\label{fig_Shenzhen}
\end{figure}

\section{Method}

\subsection{Data}
We used two datasets that include publicly available datasets from Montgomery County, Maryland, and Shenzhen No. 3 People’s Hospital in China. These datasets are maintained by the National Library of Medicine (NLM), National Institutes of Health (NIH) \cite{Candemir}. In the Montgomery County X-Ray Set, there are 138 posterior-anterior X-Rays (80 X-Rays are normal, and 58 X-Rays have a wide range of abnormalities, including effusions and miliary patterns). Shenzhen Hospital X-Ray Set have 340 normal X-Rays and 275 abnormal X-Rays showing various manifestations of tuberculosis. Figure ~\ref{fig_Motgomery} shows sample data from the Montgomery dataset and Figure ~\ref{fig_Shenzhen} shows sample data from the Shenzhen dataset.

\subsection{Proposed Network Architecture}
\begin{figure}[h]
\centering
\includegraphics[width=3.5in, height=1.05in]{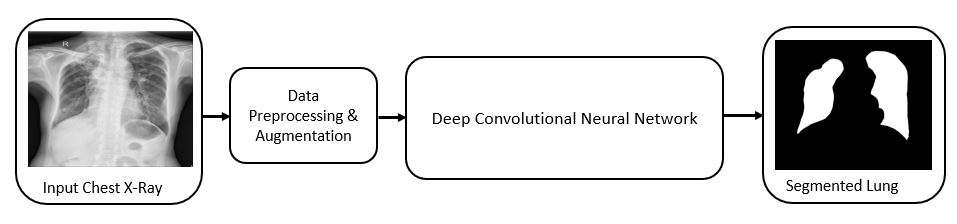}
\caption{Proposed Lung segmentation model. }
\label{fig_model}
\end{figure}

\begin{figure}[h]
\centering
\includegraphics[width=4.5in, height=4in]{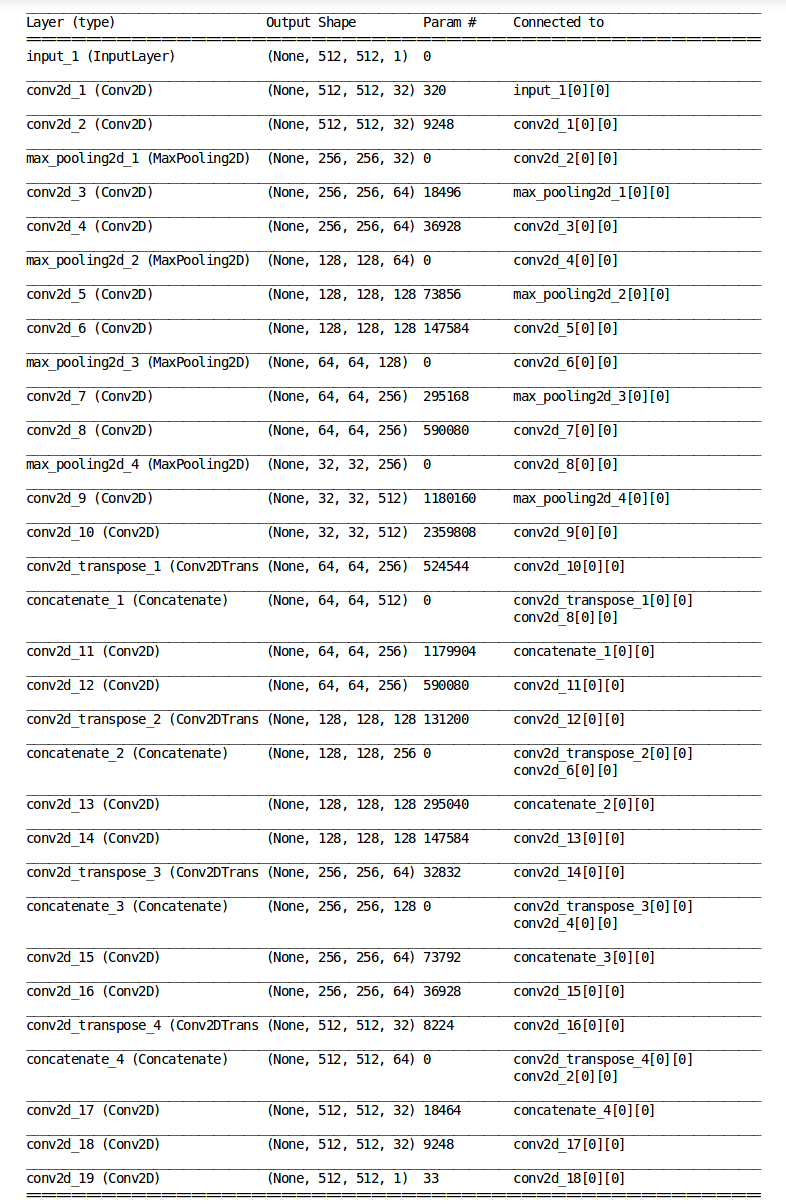}
\caption{Detail of the CNN architecture used in the proposed lung segmentation model.}
\label{fig_model_conf}
\end{figure}

Our proposed Lung segmentation model is based on the U-Net \cite{unet} architecture. The proposed model is shown at Figure ~\ref{fig_model}. Several data augmentation techniques such as zooming, cropping, horizontal flipping etc. are performed on the input dataset for increasing the training data volume. After data pre-processing and data augmentation, each image is resized to 512*512 dimension. Similar to U-Net \cite{unet}, the lung segmentation network consists of a contracting path and an expansive path. Upsampling of the feature map in the expansive path is combined with the high resolution features from the contracting path to retain the segmentation information. Detail of the CNN architecture used in the proposed lung segmentation model is shown in Figure~\ref{fig_model_conf}.

\section{Experiments}
Each chest X-ray image was resized to 512*512 before passing to the Deep CNN model for segmentation. We have combined the left and right masks for each chest X-Ray of the Montgomery dataset and performed Morphological transformations (Dilation) on the combined mask. We have developed the proposed model using Keras framework and trained on NVIDIA GTX TITAN V GPU. We have combined all the chest X-Rays from both Montgomery and Shenzhen dataset for building our input dataset. The training set consists of 80\% data of the total input dataset, and 20\% data were used as the test dataset. From the training dataset, 10\% data were used as validation dataset. We trained the proposed model using the Adam optimizer with learning rate of 0.0005, batch size of 4 and 200 epochs. For data augmentation, zoom range was set to 0.05, height shift, width shift and horizontal shift was used.

\subsection{Results}
Table~\ref{table_comparison} shows the result of our proposed Lung segmentation model. We report the result in terms of DICE coefficient \cite{dice} following previous research works. DICE coefficient is the overlap between the ground truth, GT and the calculated segmented mask, S:
\begin{equation}
DICE = \frac{|S\cap GT|}{|S| + |GT|} = \frac{2 |TP|}{2 |TP| + |FN| + |FP|}
\end{equation}
\begin{table}[h]
\centering
\caption{Performance comparison of the proposed model with previous state-of-the-arts.}
\label{table_comparison}
\begin{tabular}{|l|c|}
\hline
Method  & Dice Coefficient\\ \hline
Candemir et al. \cite{Candemir} & 94.1  \\ \hline
ED-CNN \cite{lcn} & 97.4 \\ \hline
FCN \cite{fcn}  & 97.7 \\ \hline
Proposed model  &  98.6  \\ \hline
\end{tabular}
\end{table}

\begin{figure}[h]
\centering
\includegraphics[width=5in, height=4.5in]{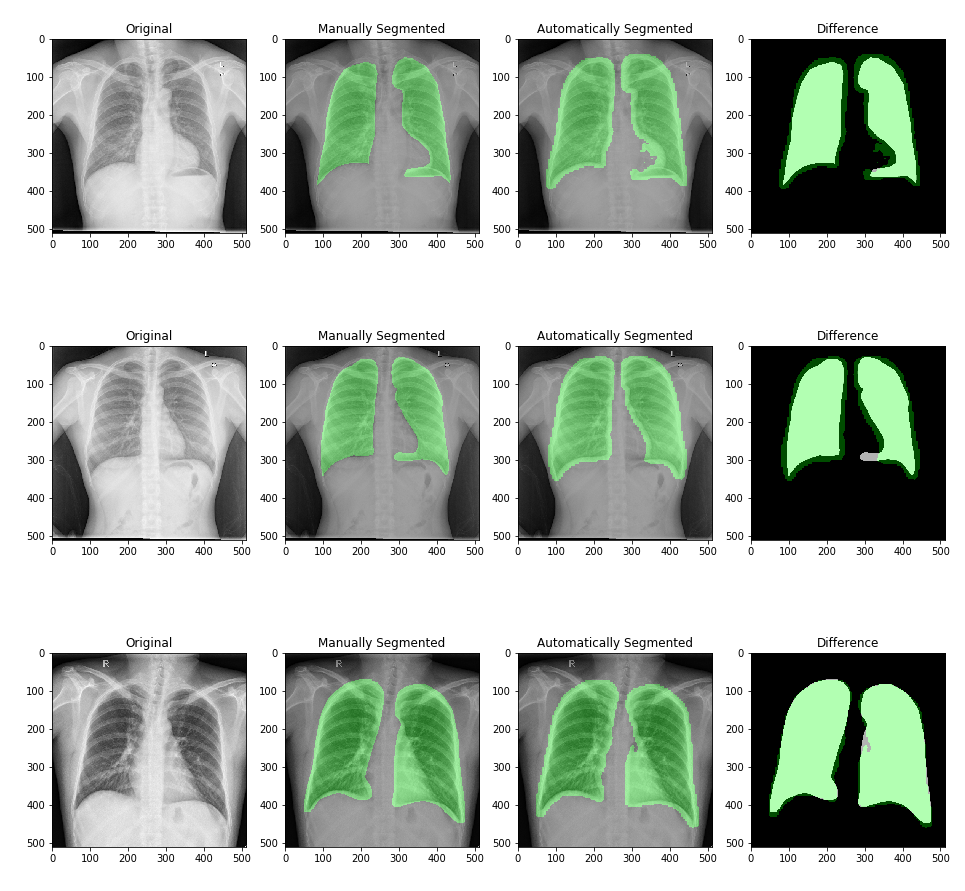}
\caption{Sample segmentation result as compared with the manual ground truth segmentation.}
\label{fig_result}
\end{figure}

A model with a higher DICE score indicates better segmentation performance of the network.
From the result, we can see that data augmentation improves the performance of the proposed network by around 2.2\%.  We have developed another CNN model using skip connections in the convolutional layers. But skip connections/resnet blocks did not help to improve the segmentation result, hence we are not reporting it here. Figure~\ref{fig_result} shows the lung segmentation result of our proposed model. The performance is consistent across different runs. From the figure, we can see that the predicted segmentation of our proposed model matches very well with the manual segmentation ground truths.

\section{Conclusion}
The proposed model demonstrates robust performance for Lung segmentation from Chest X-Rays. In future, we will evaluate the performance of our proposed model for other Chest X-Ray database including JSRT. Additionally, instead of using another mask database, we will incorporate attention mechanism in the proposed architecture for improving the segmentation result and performing weakly supervised segmentation.



\end{document}